\documentclass[lettersize,journal]{IEEEtran}
\usepackage{amsmath,amssymb,amsfonts}
\usepackage{algorithmic}
\usepackage{algorithm}
\usepackage{array}
\usepackage{graphicx}
\usepackage{textcomp}
\usepackage{stfloats}
\usepackage{booktabs}
\usepackage{subfig}
\usepackage{subfloat}
\usepackage{multirow}
\usepackage{multicol}
\usepackage{stfloats}
\usepackage{siunitx}
\usepackage{url}
\usepackage{verbatim}
\usepackage[table]{xcolor}
\usepackage{graphicx}
\usepackage{amsmath}
\usepackage{breqn}
\usepackage[algo2e]{algorithm2e}
\usepackage{cite}
\usepackage{threeparttable} 
\begin{document}

\title{Towards Stable Cross-Domain Depression Recognition under Missing Modalities}

\author{Jiuyi Chen$^{\ast}$, Mingkui Tan$^{\ast}$, Haifeng Lu, Qiuna Xu, Zhihua Wang, Runhao Zeng$^{\dagger}$, and Xiping Hu

\thanks{Jiuyi Chen is with (1) School of Future Technology, South China University of Technology, Shenzhen, 511442, Guangdong, China, and (2) PengCheng Laboratory, Shenzhen, 518057, Guangdong, China (E-mail: ftchenjiuyi@mail.scut.edu.cn).}
\thanks{Mingkui Tan is with (1) School of Software Engineering, South China University of Technology, Guangzhou, 510006, Guangdong, China, and (2) PengCheng Laboratory, Shenzhen, 518057, Guangdong, China (E-mail: mingkuitan@scut.edu.cn).}
\thanks{Haifeng Lu, Runhao Zeng, Xiping Hu are with (1) Artificial Intelligence Research Institute, Shenzhen MSU-BIT University, Shenzhen, 518172, China, and (2) Guangdong-Hong Kong-Macao Joint Laboratory for Emotional Intelligence and Pervasive Computing, Shenzhen, 518172, Guangdong, China (E-mail: luhf18@lzu.edu.cn; zengrh@smbu.edu.cn; huxp@bit.edu.cn).}
\thanks{Qiuna Xu is with School of Computer Science and Technology, Guangdong University of Technology, Guangzhou, 510006, Guangdong, China (Email: 3222004726@mail2.gdut.edu.cn).}
\thanks{Zhihua Wang is with Department of Computer Science, City University of Hong Kong, Kowloon, 999077, China (Email: zhihua.wang@my.cityu.edu.hk).}

\thanks{$^{\dagger}$ Corresponding author.}
\thanks{$^{\ast}$ Equal contribution.}
}



\maketitle

\begin{abstract}
Depression poses serious public health risks, including suicide, underscoring the urgency of timely and scalable screening. Multimodal automatic depression detection (ADD) offers a promising solution; however, widely studied audio- and video-based ADD methods lack a unified, generalizable framework for diverse depression recognition scenarios and show limited stability to missing modalities, which are common in real-world data. In this work, we propose a unified framework for \textbf{S}table \textbf{C}ross-Domain \textbf{D}epression Recognition based on \textbf{M}ultimodal \textbf{L}arge \textbf{L}anguage \textbf{M}odel (SCD-MLLM). The framework supports the integration and processing of heterogeneous depression-related data collected from varied sources while maintaining stability in the presence of incomplete modality inputs. 
Specifically, SCD-MLLM introduces two key components: (i) Multi-Source Data Input Adapter (MDIA), which employs masking mechanism and task-specific prompts to transform heterogeneous depression-related inputs into uniform token sequences, addressing inconsistency across diverse data sources; (ii) Modality-Aware Adaptive Fusion Module (MAFM), which adaptively integrates audio and visual features via a shared projection mechanism, enhancing resilience under missing modality conditions.
We conduct comprehensive experiments under multi-dataset joint training settings on five publicly available and heterogeneous depression datasets from diverse scenarios: CMDC, AVEC2014, DAIC-WOZ, DVlog, and EATD. Across both complete and partial modality settings, SCD-MLLM outperforms state-of-the-art (SOTA) models as well as leading commercial LLMs (Gemini and GPT), demonstrating superior cross-domain generalization, enhanced ability to capture multimodal cues of depression, and strong stability to missing modality cases in real-world applications.
\end{abstract}


\section{Introduction}
\label{sec:introduction}
Depression is a widespread mental disorder, with severe cases potentially leading to suicidal behavior. To enable more timely and scalableassessment beyond traditional self-report and clinical interviews, recent research has explored AI-based Automatic Depression Detection (ADD) methods, leveraging audio \cite{3deprssionsucide}, text \cite{9ADD}, facial expressions \cite{shangguan2025facial}, skeleton movements \cite{depressionbutai}, and physiological signals \cite{depressionnaodian} to build objective and efficient recognition models. Among the various modalities, audio, text, and video are the most widely used due to their accessibility and rich capacity to convey emotional cues. However, current ADD approaches with these three modalities still face two critical challenges.

\begin{table*}[!tp]
    \centering
    \caption{Overview of publicly available depression datasets. Table highlights the significant variability in data sources, modalities, and feature extraction methods. These differences underscore the challenges of building a unified model capable of generalizing across diverse sources.} 
    \label{table:contrast}
    \footnotesize
    \renewcommand{\arraystretch}{1.25}
    \setlength{\tabcolsep}{4.0mm}
\begin{tabular}{c|c|c|c|c|c|c}
\hline
\textbf{Public   Databases} & \textbf{Data Sources}            & \begin{tabular}[c]{@{}c@{}}\includegraphics[width=0.02\textwidth]{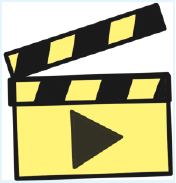} \ \textbf{Video}\end{tabular} & \begin{tabular}[c]{@{}c@{}}\includegraphics[width=0.022\textwidth]{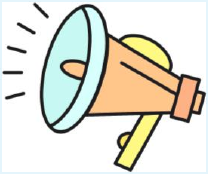} \ \textbf{Audio}\end{tabular} & \begin{tabular}[c]{@{}c@{}}\includegraphics[width=0.017\textwidth]{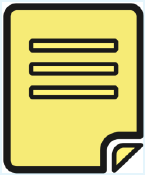} \ \textbf{Text}\end{tabular} & \textbf{Video   Features}                                                     & \textbf{Audio   Features}                                                  \\ \hline
\textbf{CMDC   \cite{CMDC}}               & Interview            & $\times$          & $\checkmark$         & $\checkmark$  & \begin{tabular}[c]{@{}c@{}}OpenFace\\/TimesFomer\end{tabular} & \begin{tabular}[c]{@{}c@{}}OpenSmile\\/VGGish \end{tabular} \\ \hline
\textbf{AVEC   2014 \cite{AVEC2014}}          & Interview            & $\checkmark$         & $\checkmark$         & $\checkmark$  & $\times$                                                                    & $\times$                                                                 \\ \hline
\textbf{DAIC  \cite{DAIC} }          & Interview            & $\times$          & $\checkmark$         & $\checkmark$  & \begin{tabular}[c]{@{}c@{}}OpenFace\\/CNN-Resnet\end{tabular} & Covarep                                                          \\ \hline
\textbf{D-Vlog   \cite{DVlog} }            & Self-narrated Video  & $\checkmark$         & $\checkmark$         & $\checkmark$  & DLIB                                                                 & OpenSmile                                                         \\ \hline
\textbf{LMVD   \cite{LMVD}}               & Self-narrated Video  & $\times$          & $\times$          & $\times$   & OpenFace                                                             & VGGish                                                            \\ \hline
\textbf{EATD   \cite{EATD} }              & Online Questionnaire & $\times$          & $\checkmark$         & $\checkmark$  & $\times$                                                                    & $\times$                                                                 \\ \hline
\end{tabular}
\begin{flushleft}
{\footnotesize Note: The symbol $\times$ indicates unavailable, while $\checkmark$ denotes available.}
\end{flushleft}
\end{table*}

\textbf{First}, current depression recognition models are still limited in their ability to provide a unified, generalizable framework capable of handling heterogeneous data across diverse recognition scenarios. This is primarily due to differences in data sources, including clinical interviews, online questionnaires, and self-narrative recordings, inconsistent modalities, and varying feature extraction strategies, as shown in Table~\ref{table:contrast}. As a result, models trained on one dataset scenario often perform poorly when transferred to another. For example, the method proposed by Chen et al. \cite{IIFDD} achieved excellent results on CMDC interview dataset but experienced a 45\% drop in performance when applied to the structurally different EATD questionnaire dataset.

\textbf{Second}, existing depression recognition models struggle to adapt to scenarios with missing modalities. This is because they rely on complete and consistent data for training and inference. As a result, when incomplete samples are used for prediction, model performance significantly degrades or even fails. However, missing modalities are common in depression recognition, particularly in real-world scenarios. For example, AVEC2014 dataset lacks audio and transcripts in some samples, and DVlog dataset suffers from missing facial recordings due to camera misalignment. This underscores the critical need for stability in real-world applications.

Overall, current multimodal depression recognition methods are limited by data heterogeneity, missing modalities, and cross-dataset feature inconsistencies due to diverse scenarios. These challenges highlight the urgent need for a unified framework that can adapt to diverse data sources while remaining stable to incomplete modalities, which is crucial for real-world deployment. However, building such a model remains challenging. It must jointly model multiple modalities, handle variable-length inputs, and deliver consistent, accurate emotion recognition despite partial modality absence.

Inspired by the powerful capabilities of large language models (LLMs) in contextual understanding, and few-shot generalization \cite{motivationllms,motivationllms2}, we explore LLMs as backbone models for multimodal depression recognition because LLMs naturally can encode rich semantic cues in depression-related text and provide an interface into which audio and visual representations can be integrated. 

However, building a unified LLM-based framework with strong generalization and stability raises three key challenges: 

\textbf{1) Difficulty in multimodal alignment}. The gap between multimodal depression data and the LLM language space requires effective alignment to enable the LLM to comprehend depression-related semantic information.

\textbf{2) Variability in modality structure and length}. The significant differences in modality structure and length across depression datasets with varying data sources hinder the ability of large models to process inputs effectively.

\textbf{3) Instability under missing modalities}. The prevalence of missing modalities in depression datasets leads to unstable training dynamics and inconsistent predictions, undermining the reliability of LLM-based multimodal depression recognition.

In this study, we propose \textbf{SCD-MLLM}, an LLM-based multimodal depression recognition framework that can align heterogeneous multimodal depression signals with the language space of the backbone LLM for \textit{Challenge 1}. It also can enable unified processing of depression-related data from diverse sources, such as interviews, questionnaires, and self-reports, while maintaining stability in the presence of missing modalities. Specifically, it introduces two core components: (i) \textbf{Multi-Source Data Input Adapter (MDIA)}, which targets \textit{Challenge 2} by using a masking mechanism with task-specific prompts to standardize heterogeneous, variable-length textual inputs into unified token sequences, addressing generalization issues across diverse data formats. (ii) \textbf{Modality-Aware Adaptive Fusion Module (MAFM)}, which targets \textit{Challenge 3} by detecting the availability of audio and visual inputs and selects different processing paths accordingly. It performs modality fusion and projects the features into the language space via a shared linear layer when both audio and visual inputs are present. If one modality is missing, it directly projects the available modality through the same layer. Extensive experiments on five publicly available and heterogeneous depression datasets from various scenarios show that SCD-MLLM outperforms state-of-the-art (SOTA) methods and leading commercial LLMs GPT-4o \cite{hurst2024gpt} and Gemini 2.5 \cite{team2024gemini}, across various modality configurations, demonstrating stable performance even with missing modalities.

In summery, our contributions include three aspects:
\begin{itemize}
    \item We propose SCD-MLLM, a unified multimodal depression recognition framework based on LLM. It supports cross-dataset training and inference, enabling generalized recognition across diverse scenarios, while maintaining stability to missing modalities.
    \item We design Multi-Source Data Input Adapter (MDIA), which leverages scenario-specific prompts and masking mechanism to align heterogeneous and variable-length inputs, addressing inconsistency across diverse data sources.
    \item We introduce Modality-Aware Adaptive Fusion Module (MAFM), which employs a modality-aware gating mechanism designed for MLLM that detects which modalities are present and adaptively routes them through appropriate processing paths, enabling LLM-based framework to handle both complete and incomplete inputs.
\end{itemize}

This paper is organized as fol1ows: Section \ref{sec:relatedworks} provides a review of prior research on multimodal depression recognition and MLLMs. 
The proposed SCD-MLLM model is presented in Section \ref{sec:methodology}. Section \ref{Experimental set} and Section \ref{sec:results_analysis} describe the dataset employed and preprocessed in this study and reports the experimental results. Finally, Section \ref{conclusion} concludes the paper.

\section{Related Work}
\label{sec:relatedworks}
In this section, we review existing multimodal depression recognition and multimodal large language models approaches and summarize current challenges in the field of depression recognition.

\subsection{Multimodal Depression Recognition}
Multimodal depression recognition leverages complementary information across modalities to improve performance \cite{MultimodalRelatedwork}. However, due to privacy constraints, many datasets release only pre-extracted facial cues, such as facial landmarks, instead of raw video. Consequently, recent work focuses on visual cue–based methods, which is also the focus of this study.

Fu et al. \cite{fu2025facial} used facial action units to derive emotion-relevant, identity-invariant descriptors. Tao et al. \cite{tao2024depmstat} encoded landmarks and audio into spatio-temporal tokens and fused them via a dedicated STAT module. Yang et al. \cite{yang2025spike} combined facial landmarks with audio using a spike-based attention mechanism, and Li et al. \cite{jung2024hique} leveraged hierarchical question embeddings and cross-modal attention over audio, text and visual cues. In contrast, our method jointly encodes multiple facial cues (landmarks, action units, gaze, and head pose) within a unified video encoder, yielding richer and more consistent visual representations for cross-dataset multimodal fusion.

Despite notable advances in depression recognition, many existing methods are confined to specific datasets and exhibit poor generalization and limited stability, particularly under missing modality conditions. 

\subsection{Large Language Model}
Large language models (LLMs) have achieved remarkable success across various domains. To further enhance their performance and generalization, recent research has explored integrating multimodal data into LLM training. In emotion understanding, multimodal LLMs, which incorporate visual, auditory, and linguistic cues, can more accurately capture emotional dynamics. As a result, leveraging LLMs for depression recognition has emerged as a growing focus within affective computing.

\subsubsection{Emotion-Aware Large Language Model}
A growing line of research explores emotion-aware multimodal LLMs. Omni-Emotion \cite{yang2025omni} captures subtle facial micro-expressions and auditory cues through a multimodal LLM framework. Emotion-LLaMA \cite{cheng2024emotion} enhances multimodal emotion understanding by integrating audio, visual, and textual inputs, further boosted by instruction tuning. EmoLLM \cite{yang2024emollm} targets emotionally supportive dialogue generation for therapeutic use. EAI-LLM \cite{lu2025understanding} performs emotion recognition from 3D skeletal data and generates fine-grained emotional descriptions. AffectGPT \cite{lian2025affectgpt} introduces a pre-fusion strategy to better integrate audio and visual modalities for more nuanced emotion analysis.

Inspired by AffectGPT's pre-fusion concept, we propose a novel adaptive multimodal fusion mechanism tailored for depression recognition, a task often constrained by limited data and missing modalities. Unlike the use of separate linear layers of AffectGPT for each modality, our method adopts a shared linear layer for audio and video inputs, enabling more flexible and robust fusion under incomplete modality conditions.

\subsubsection{Depression-Aware Large Language Model}
LLM-based depression recognition remains relatively underexplored. Zhang et al. \cite{zhang2024llms} injected acoustic tokens into a LLM to analyze depression-related speech patterns, but their approach is restricted to the audio modality and evaluated only on DAIC dataset. Wang et al. \cite{zhang2025mllmdrexplainabledepressionrecognition} proposed a multimodal LLM for explainable depression diagnosis in interview videos, using a lightweight query module and smaller fine-tuned LLMs, yet it remains confined to interview-style data without cross-scenario validation. 

In contrast, our approach supports cross-scenario depression recognition and introduces a novel video encoder for extracting depression-related visual features. By integrating text and audio modalities through multimodal fusion, our method enhances both the accuracy and stability of depression recognition.

\section{Methodology}
\label{sec:methodology}

\textbf{Notion.} Let $D_{i}^{j}({t,a,v^{(*)}})$ denote the $i$-th depression dataset ($i=1,\dots,N$) of scenario $j \in \left \{ I, Q, S \right \}$, where $t$, $a$, and $v$ refer to the text, audio, and video modalities with corresponding sequence lengths, respectively. Here, $I$, $Q$, and $S$ indicate \textbf{I}nterview, \textbf{Q}uestionnaire, and \textbf{S}elf-narration types, and the superscript $(*)$ denotes missing modalities.

\textbf{Problem Definition.} Models trained on a specific dataset $D^{j}({t,a,v})$ often fail to generalize to other datasets of differert scenarios $D^{k}(t,a,v^{(*)})$ with different modality configurations and sequence length distributions. As shown in Table \ref{table:contrast}, this is due to different $j$ correspond to different data acquisition protocols, leading to diverse modality length. Meanwhile, some datasets release raw videos, others provide only extracted visual features using heterogeneous tools because of privacy concerns, causing inconsistency across video modality, and even for the same data source, such as $j=S$, feautre extraction tools vary, making their $a$ and $v$ incompatible.

At the same time, existing models are suitable only for training and inference where all modalities are available, but many real-world depression scenarios always contain missing modalities. For example, due to visual privacy concerns, some datasets present missing video modalities, denoted as $D({t,a,v^{({*})}})$; others may suffer from missing audio and text $D({t^{(*)},a^{(*)},v})$ due to recording failures.

These challenges highlight the urgent need for a unified framework that can adapt to diverse data sources with varying modality configurations and sequence lengths while remaining robust to incomplete modalities, which is crucial for real-world deployment.

\begin{figure*}[!t]
\centering
\includegraphics[width=1.0\textwidth]{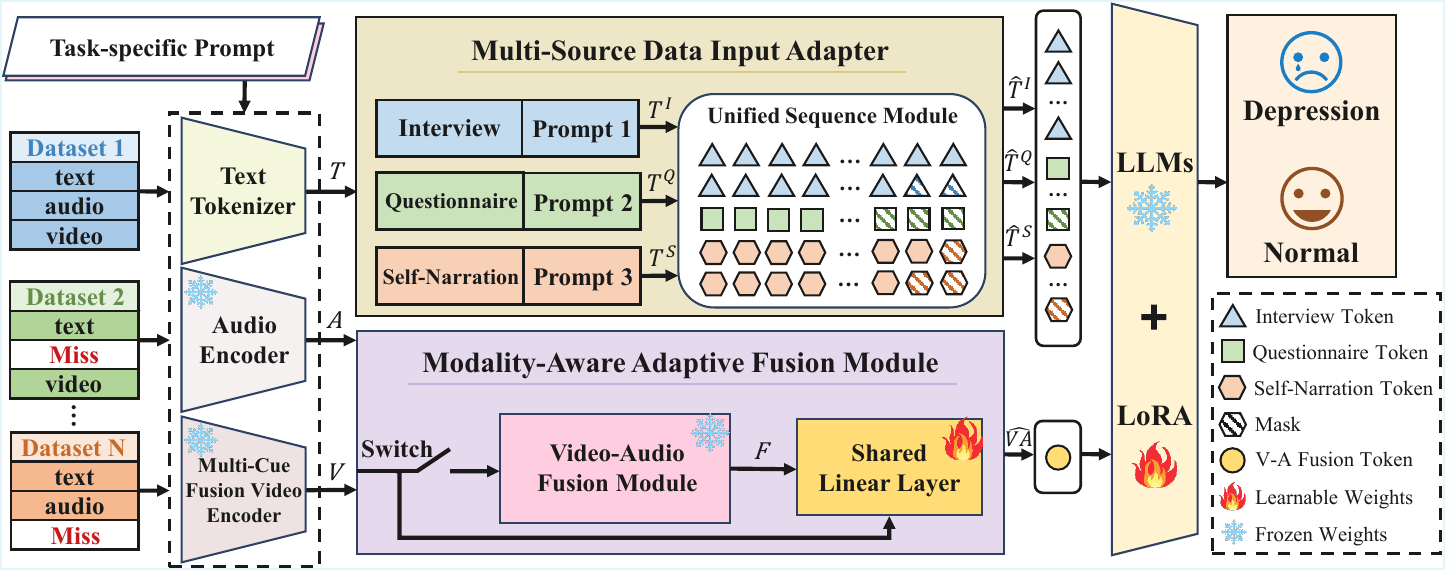} 
\caption{Overview of the proposed SCD-MLLM framework. This work presents a unified cross-domain multimodal depression recognition framework with strong stability based on large language model (LLM), termed SCD-MLLM. Unlike prior approaches, SCD-MLLM transforms heterogeneous depression-related inputs into uniform token sequences through a Multi-Source Data Input Adapter (MDIA), and employs a Modality-Aware Adaptive Fusion Module (MAFM) to adaptively integrate audio-visual cues, enabling stable inference under incomplete modality conditions.} 
\label{fig:overview-architecture}
\vspace{-0.2cm}
\end{figure*}

\subsection{General Scheme}
To address aforementioned challenges, we aim to leverage LLM to develop a unified cross-domain multimodal depression recognition framework $F(\cdot \ ,\cdot \ ,\cdot)$, as shown in Fig. \ref{fig:overview-architecture}. This framework enables depression recognition results $O$ to be obtained even in the presence of missing modalities or varying data collection scenarios across depression datasets, as demonstrated in Eq. (\ref{eq:overview}).
\begin{multline}
O = F(D_{1}^{j}({T^{(*)},A^{(*)},V}), D_{2}^{j}({T,A,V^{(*)}}), \ldots, \\
      D_{N}^{j}({T,A,V})), \quad j \in \{ I, Q, S \}
\label{eq:overview}
\end{multline}

We use LLM as the backbone because it natively operates on text, which often carries rich semantic indicators of depressive symptoms, and can capture subtle contextual and affective nuances in natural language. Moreover, its strong few-shot generalization and token-based architecture provide both the potential to build a generalizable depression recognition model and a natural interface for integrating additional modalities.

However, building a unified LLM with strong generalization and robustness faces three challenges. 

\textbf{C1: Difficulty in multimodal alignment}. Multimodal depression signals are difficult to align with the LLM language space (token level), limiting the model’s ability to capture depression-related semantics.

\textbf{C2: Variability in modality structure and length}. Pronounced cross-domin variation in modality structure and sequence length makes it difficult for LLM to process inputs consistently.

\textbf{C3: Instability under missing modalities}. The prevalence of missing modalities complicates both training and inference.

To address \textbf{C1}, each modality is processed through its respective encoder to extract features and generate unified representations, as detailed in Sec.~\ref{sec:Multiencoder}. This ensures a shared foundation for downstream modeling and fusion. For \textbf{C2}, we propose \textbf{Multi-Source Data Input Adapter (MDIA)}, which normalizes heterogeneous, variable-length textual inputs into unified token sequences using masking mechanism and task-specific prompts. This design standardizes the input data, improving model performance across different datasets obtained from diverse depression recognition scenarios. To solve \textbf{C3}, we introduce \textbf{Modality-Aware Adaptive Fusion Module (MAFM)} in Sec.~\ref{sec:MAFM}, which adaptively integrates audio and visual features. When both modalities are available, it performs fusion and projects the fused features into the language space via a shared linear transformation. If one modality is missing, it directly projects the available modality, ensuring consistent semantic alignment and enhancing stability under incomplete data 

\subsection{Multimodal Feature Extraction}
\label{sec:Multiencoder}
In this section, we encode each modality into token-level embeddings and describe the design of the corresponding encoders to enable LLM-based multimodal alignment and understanding.
\subsubsection{Text Embed with Tokenizer}
We use the tokenizer and embedding layer of the LLM to convert the task-specific prompt $p$ and raw text $t$ into a sequence of token embeddings:
\begin{equation}
\begin{aligned}
T^{I} &= Tokenizer(p^I, t^I)\in\mathbb{R}^{l_I\times d_{llm}},\\
T^{Q} &= Tokenizer(p^Q, t^Q)\in\mathbb{R}^{l_Q\times d_{llm}},\\
T^{S} &= Tokenizer(p^S, t^S)\in\mathbb{R}^{l_S\times d_{llm}},
\end{aligned}
\label{eq:text_Emb}
\end{equation}
where \(l_I\), \(l_Q\) and \(l_S\) is the text token sequence length of different input and $d_{llm}$ denotes the hidden size (embedding dimension) of the LLM. The sequence $T$ is then fed into the multimodal framework for further processing.

\subsubsection{Audio Embedding Extraction}
To capture prosodic features linked to depressive emotions, such as rhythm, intonation, and speech rate, we design an audio encoding module based on Mel-spectrograms and NetVLAD aggregation \cite{arandjelovic2016netvlad}. This module summarizes frame-level acoustic patterns into compact token-level embeddings, facilitating consistent alignment with other modalities for multimodal fusion.

Given a waveform audio input, we compute its log-Mel spectrogram using a set of Mel filter banks, preserving perceptual and temporal features. NetVLAD then applies a soft-assignment mechanism, similar to K-means clustering, to derive a global representation from local frame features $\mathbf{a}_i$, as shown in Eq. (\ref{eq:netvlad}).
\begin{equation}
\begin{gathered}
\mathrm{vlad}[k] = \sum_{i=1}^t \alpha_k(i)\, (\mathbf{a}_i - \mathbf{c}_k), \quad k=1, \dots, K \\
A = \mathcal{L}_2(\mathrm{vlad}[1], \mathrm{vlad}[2], \dots, \mathrm{vlad}[K]),
\end{gathered}
\label{eq:netvlad}
\end{equation}
where $t$ is the frame length, $K$ is the number of cluster centers, $\alpha_k(i)$ is the soft-assignment weight for the $i$-th frame to the $k$-th cluster center $\mathbf{c}_k$, and $\mathcal{L}_2$ denotes intra-cluster L2 normalization followed by global L2 normalization. With $K$ cluster centers, the resulting audio embedding $A$ is then fed into our multimodal framework for further processing.

\subsubsection{Multi-Cue Fusion Video Encoder}
\label{sec:MFVE}
\begin{figure}[!t]
\centering
\includegraphics[width=0.48\textwidth]{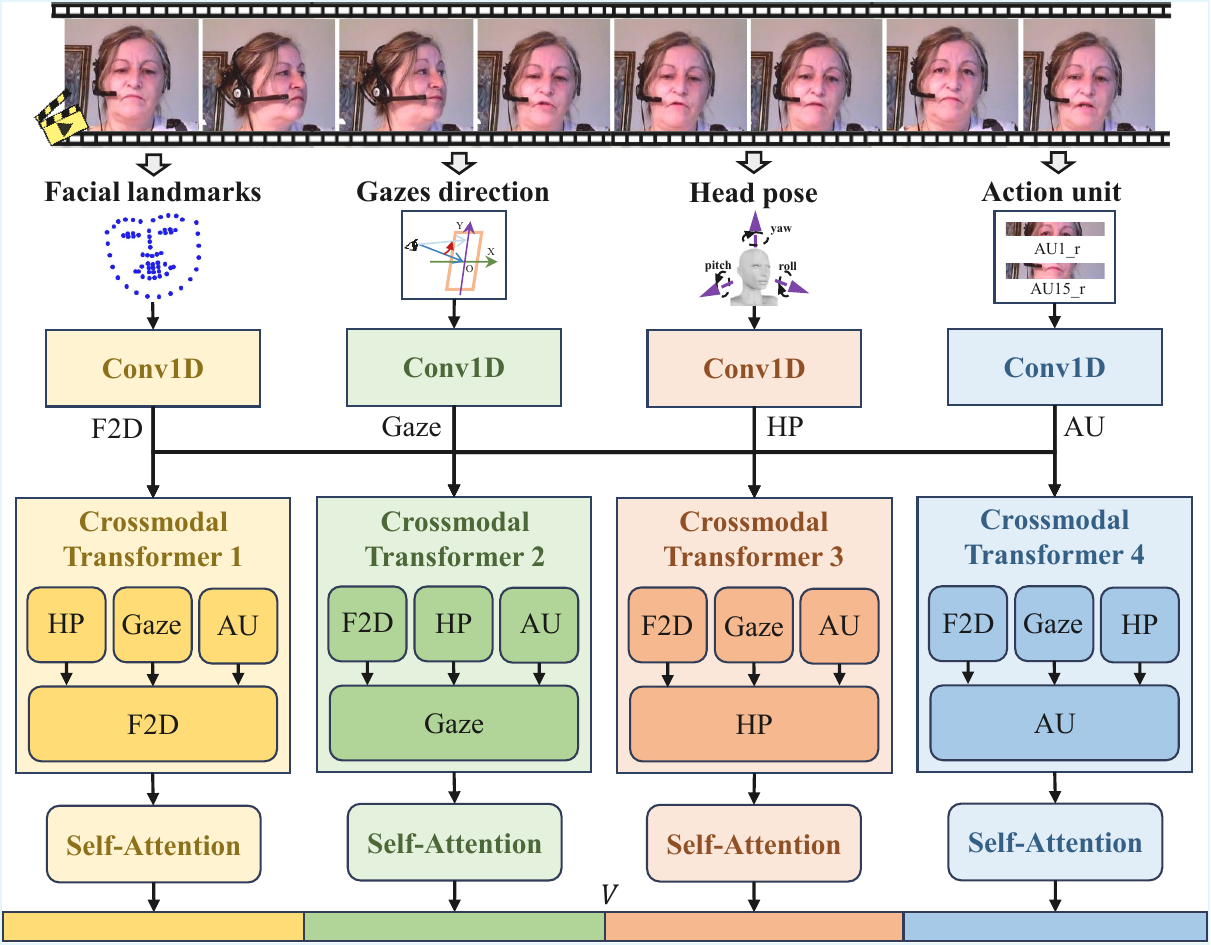} 
\caption{The framework of Multi-Cue Fusion video Encoder (MFVE). This module captures the high-order interactions among heterogeneous facial dynamics through crossmodal and self-attention mechanisms, enabling semantically aligned and LLM-compatible video representations for depression understanding.} 
\label{fig:MFVE}
\vspace{-0.2cm}
\end{figure}

In the context of depression recognition, facial video data contain rich emotional cues that are crucial for effective emotion modeling. However, due to privacy constraints, most public datasets provide only pre-extracted visual features instead of raw videos (Table~\ref{table:contrast}). A comparative analysis shows that many datasets release facial features extracted with OpenFace, whereas a few, such as DVlog, provide only raw videos, from which we can extract OpenFace-based features ourselves. To ensure cross-dataset consistency, we therefore uniformly use OpenFace \cite{openface2.0} to obtain four standardized facial cues: facial landmarks (F2D), eye gaze direction (Gaze), head pose (HP), and facial action units (AU), each carrying semantically relevant emotional information \cite{wang2025facial}. Nonetheless, their heterogeneous dimensionalities and temporal structures hinder unified visual modeling, which is essential for capturing joint spatiotemporal dependencies and complementary depression-related semantics.

To address this issue, we introduce the \textbf{Multi-Cue Fusion Video Encoder} (MFVE), which integrates all cues into a unified visual representation (Fig.~\ref{fig:MFVE}). To regularize the structure of multi-cue visual features and enhance their semantic consistency, we first apply a one-dimensional convolution (Conv1D) to each cue sequence $F^m$ to extract local temporal features $\widetilde{F}^m$:
\begin{equation}
\widetilde{F}^m = \mathrm{Conv1D}(F^m),\ m\in\{\mathrm{F2D},\mathrm{Gaze},\mathrm{HP},\mathrm{AU}\}.
\label{eq:mfve_conv}
\end{equation}

Next, a Crossmodal Transformer is employed to capture contextual relationships across cues and refine emotional representations. Specifically, it models inter-cue dependencies via cross-attention, where the current cue $\widetilde{F}^m$ serves as the query $Q^m$ and the remaining cues $\widetilde{F}^n$ serve as keys $K^n$ and values $V^n$:
(\ref{eq:mfve_crossmodal}).
\begin{equation}
\label{eq:mfve_crossmodal}
\begin{aligned}
&\mathrm{Attention}^{m,n} =
    \mathrm{Softmax}\!\left(\frac{Q^m (K^n)^\top}{\sqrt{d_k}}\right)V^n, \\
&\widehat{F}^m =
    \mathrm{LayerNorm}\!\left(\widetilde{F}^m +
    \sum_{n\neq m} \mathrm{Attention}^{m,n}\right),
\end{aligned}
\end{equation}
where \(d_k\) denotes the dimensionality of the key vectors.

Finally, the four cross-enhanced representations are aggregated using a Self-Attention ($SA$) module and concatenated into a unified visual token embedding $V$:
\begin{equation}
V=\left[SA(\widehat{F}^{\mathrm{F2D}});SA(\widehat{F}^{\mathrm{Gaze}});SA(\widehat{F}^{\mathrm{HP}});SA(\widehat{F}^{\mathrm{AU}})\right].
\label{eq:mfve_selfattention}
\end{equation}

MFVE both standardizes video representations across heterogeneous datasets and enhances the expressiveness of the visual modality, thereby improving the ability of the LLM to exploit facial cues for multimodal depression recognition.

\subsection{Multi‑Source Data Input Adapter}
\label{sec:MDIA}
Multimodal depression datasets come from various sources, such as structured interviews, dynamic questionnaires, and self-narratives, each with different text structures, modalities, and sequence lengths. These differences pose challenges for unified modeling, often requiring separate models for each dataset and leading to poor cross-dataset generalization. To address this, we propose the \textbf{Multi-Source Data Input Adapter (MDIA)}, which uses task-specific prompts and a masking mechanism to standardize input length and structure, ensuring consistent token representations for subsequent multimodal fusion.

\subsubsection{Prompt Design and Data Source Awareness} 
\label{sec:promptset}
We design tailored prompts for each data source: Interview prompts focus on question-answer alignment, questionnaire prompts guide emotion understanding across varied items, and self-narrative prompts extract key emotional cues from free-form text (see Fig.~\ref{fig:promptset}). This approach preserves semantic integrity and enhances contextual understanding, enabling the LLM to generate consistent and accurate embeddings across different input types.

\subsubsection{Token Embedding Sequence Unification}
To ensure effective cross-scenario learning, we propose the Unified Sequence Module (USM), which standardizes token sequences from diverse inputs. Using a masking mechanism (Eq. \ref{eq:Mask}), it adjusts sequences of varying lengths into a unified fixed-length sequence, whether structured text from interviews, varied responses from questionnaires, or free-form text from self-narratives, laying the foundation for multimodal fusion with audio and video data.
\begin{equation}
    \widehat{T}^{I} = T^{I}\odot M^{I},\;
    \widehat{T}^{Q} = T^{Q}\odot M^{Q},\;
    \widehat{T}^{S} = T^{S}\odot M^{S},
\label{eq:Mask}
\end{equation}
where $T^I$, $T^Q$, and $T^S$ are the original token embeddings for the interview, questionnaire, and self-narrative modalities, and $\widehat{T}^I$, $\widehat{T}^Q$, and $\widehat{T}^S$ are the masked token embeddings. $M^{I}$, $M^{Q}$ and $M^{S}$ are a $1 \times l$ attention maps that preserve original tokens and nullify padded elements, with $l$ being the maximum token length across all inputs and $\odot$ denotes element-wise multiplication.

Overall, MDIA standardizes input lengths and structures across heterogeneous multimodal depression datasets, improving model generalization and enabling more accurate recognition across diverse real-world scenarios.

\subsection{Adaptive Video-Audio Fusion with Modality Awareness}
\label{sec:MAFM}
In multimodal depression recognition, audio and visual modalities provide key emotional cues, such as speech rate, intonation, facial expressions, and action units. However, real-world depression datasets often suffer from inconsistent modality composition and missing modalities. For instance, the EATD dataset \cite{EATD} includes only text and audio, while the CMDC dataset \cite{CMDC} has incomplete interview responses. Traditional multimodal fusion methods assume the presence of all modalities, which undermines robustness when data is incomplete.

To address this, we propose the \textbf{Modality-Aware Adaptive Fusion Module (MAFM)}, which adapts to missing modalities by performing fusion when both audio and visual modalities are available, and bypassing fusion when one is missing. This approach enhances stability under incomplete input conditions. The architecture of MAFM is shown in Fig.~\ref{fig:overview-architecture}.

\subsubsection{Adaptive Switch Mechanism for Modality Awareness}
We design an adaptive selection mechanism that checks for the presence of audio and visual inputs and selects the appropriate processing path $P(\cdot, \cdot)$. When both modalities are available, fusion is performed through the Video-Audio Fusion Module (VAFM). If one modality is missing, it bypasses fusion and directly projects the available modality to preserve depression-related semantic information. If both modalities are missing, only the text sequence is processed. This process is shown in Eq. (\ref{eq:AdaptiveSelection}).
\begin{equation}
P(A, V)=
\begin{cases}
P_{F}(A, V),\ \text{if} \ \ A\neq\emptyset \ \ \text{and} \ \ V\neq\emptyset\\
P_{B}(A, V),\ \text{if} \ \ A=\emptyset \ \ \text{or} \ \ V=\emptyset
\end{cases},
\label{eq:AdaptiveSelection}
\end{equation}
where $P_{F}(\cdot\ , \cdot)$ is fusion path and $P_{B}(\cdot\ , \cdot)$ is the bypassing path. $A$ and $V$ are the audio and video token embeddings obtained by the respective encoder in Sec.~\ref{sec:Multiencoder} and $\emptyset$ indicate empty vector for missing modality data.

\subsubsection{Video-Audio Emotional Interaction and Fusion}
To maximize the retention of meaningful information and effectively align semantic between audio and video, we introduce a cross-modal fusion structure using multi-head cross-attention to model the relationships between audio and video modalities. The audio embedding $A$ and video embedding $V$ are first processed through modality-specific fully connected layers to generate query ($Q_a, Q_v$), key ($K_a, K_v$), and value ($V_a, V_v$) representations. These representations are then fed into the attention mechanism, which captures the dynamic interactions between the two modalities, as shown in Eq. (\ref{eq:MHA}).

\begin{figure}[!t]
\centering
\includegraphics[width=0.48\textwidth]{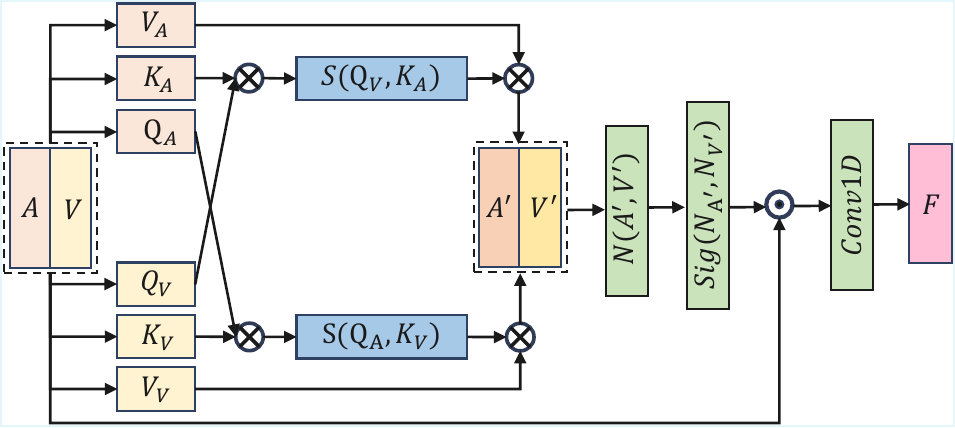} 
\caption{The framework of the Video-Audio Fusion Module (VAFM). This module captures complementary cues through cross-modal attention and fuses audio-visual features into an unified representation for downstream LLM-based depression analysis.} 
\label{fig:MAFM}
\vspace{-0.2cm}
\end{figure}

\begin{equation}
\begin{aligned}
A^{\prime}(Q_v,K_a,V_a)=\sum_{i=1}^H{S}\left(\frac{Q_vK_a^T}{\sqrt{d_K}}\right)V_a ,\\
V^{\prime}(Q_a,K_v,V_v)=\sum_{i=1}^H{S}\left(\frac{Q_aK_v^T}{\sqrt{d_K}}\right)V_v ,
\end{aligned}
\label{eq:MHA}
\end{equation}
where $A^{\prime}$ and $V^{\prime}$ are the interactive audio and video embeddings, $H$ is the number of attention heads, $S(\cdot\ , \cdot)$ denotes the softmax operation, and $d_K$ is the dimensionality of the key vectors.

To reduce modality discrepancies, the embeddings are normalized and passed through a sigmoid transformation for nonlinear feature extraction. The resulting vectors are then element-wise multiplied with the original input features, similar to residual connections, which enhances emotional cues and suppresses redundancy. Finally, the fused features are integrated using a 1D convolution ($Conv1D$), yielding a unified merge representation. This process is detailed in Eq. (\ref{eq:Sig_Normalization}).
\begin{equation}
F=Conv1D(Sig(N(A^{\prime})) \odot A+Sig(N(V^{\prime})) \odot V),
\label{eq:Sig_Normalization}
\end{equation}
where $F$ represents the fused features, $N(\cdot\ , \cdot)$ is normalization, $Sig(\cdot\ , \cdot)$ is the sigmoid function, and $\odot$ denotes element-wise multiplication.

This fusion module enhances emotional expression by integrating audio and visual modalities, capturing complementary emotional cues. By aligning their emotional semantics, the module improves the ability of LLM to more intricately understand depressive states, ensuring that both modalities contribute meaningfully to depression recognition, thereby improving the overall performance in detecting depression.

\subsubsection{Cross-Modal Shared Projection}
Whether using fused features from the full path $P_F(\cdot, \cdot)$ or single-modality features from the bypass path $P_B(\cdot, \cdot)$, all outputs are passed through a shared linear layer to project them into the LLM-compatible semantic space. This ensures semantic alignment with textual inputs and smooth integration into the unified token embedding sequence generated by MDIA. Unlike traditional methods that use separate projections for each modality, we employ a shared projection for two reasons:
\begin{itemize}
    \item Audio and visual signals in depression recognition often represent the same latent emotional state, and shared mapping captures these cross-modal emotional semantics without feature space fragmentation.
    \item A unified projection ensures consistency across diverse scenarios and facilitates seamless multimodal integration into a common semantic space.
\end{itemize}

Experimental results confirm that this design improves performance stability and model generalization. The shared linear mapping layer is illustrated in Eq. (\ref{eq:SharedLinear}):
\begin{equation}
\widehat{VA}=
\begin{cases}
\mathbf{W}\cdot F,\ \text{if} \ \ A\neq\emptyset \ \ \text{and} \ \ V\neq\emptyset\\
\mathbf{W}\cdot A,\ \text{if} \ \ A\neq\emptyset \ \ \text{and} \ \ V=\emptyset \\
\mathbf{W}\cdot V,\ \text{if} \ \ A=\emptyset \ \ \text{and} \ \ V\neq\emptyset
\end{cases}\ ,
\label{eq:SharedLinear}
\end{equation}
where $\widehat{VA}$ represents the fused video-audio token embedding, and $\mathbf{W}$ denotes the learnable weights of the shared linear layer, which adaptively adjusts the mapping between audio-video and text features during training.

MAFM addresses the challenge of missing modalities by dynamically adjusting the processing path based on modality availability. It not only enhances multimodal fusion when both audio and video are present but also ensures model stability and robustness in scenarios with missing modalities, improving performance and generalization across diverse depression recognition tasks.

\subsection{Training Multimodal Depression Large Language Model}
\label{MMDLLM}

\begin{figure*}[!t]
\centering
\includegraphics[width=\textwidth]{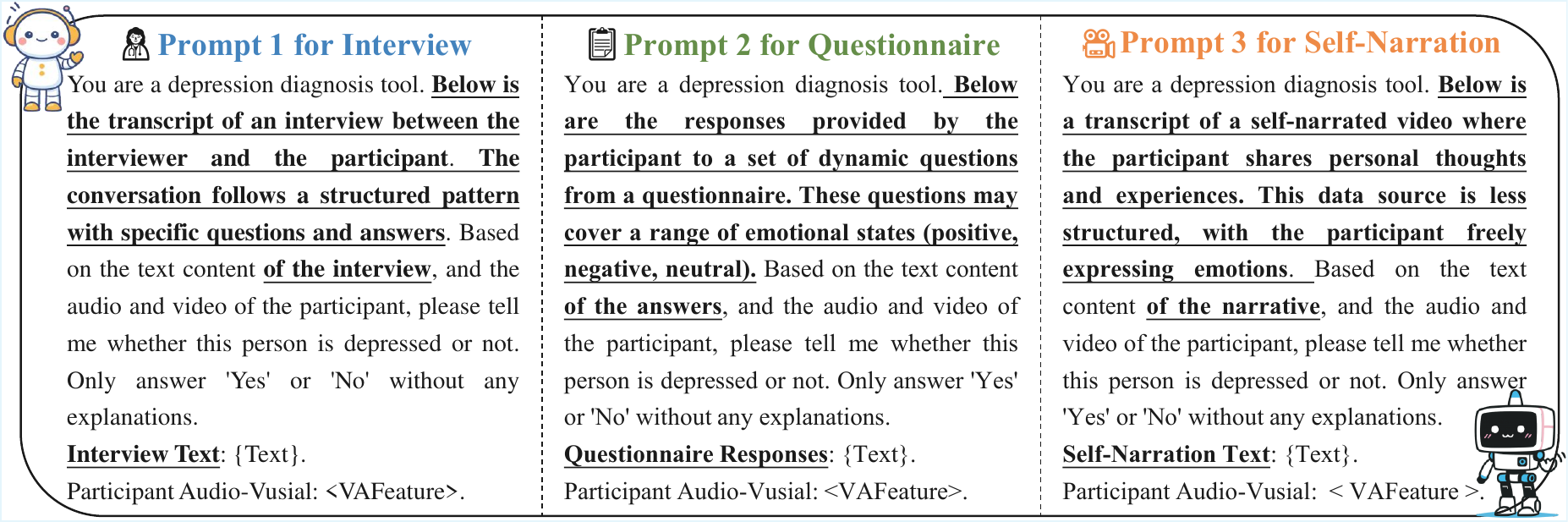}
\caption{Different prompt designs for different data sources.} 
\label{fig:promptset}
\vspace{-0.2cm}
\end{figure*}

Large Language Models (LLMs) are primarily designed for natural language processing and are not natively compatible with raw audio signals or multi-cue visual features in depression datasets. To address this, we integrate Low-Rank Adaptation (LoRA) \cite{hu2022lora} into our framework, enhancing the LLM's ability to capture depression-related affective semantics. The training process is as follows:

Step 1: Pretrain the audio encoder, video encoder (MFVE), and Video-Audio Fusion Module (VAFM) on paired audio–video samples from all datasets. The fused embeddings produced by VAFM are then passed through a classifier and optimized using a supervised depression prediction loss.

Step 2: Tokenize the text data using the LLM tokenizer, while the pre-trained audio and video encoders transform raw audio and multi-cue visual inputs into feature embeddings.

Step 3: Fuse the audio and visual features with the pre-trained VAFM and project them into a token representation that is compatible with the LLM's semantic space using a shared linear layer, bridging the modality gap.

Step 4: Design task-specific prompts for different data sources (interviews, questionnaires, and self-narrations), ensuring consistent alignment across modalities. The fused audio-visual token is inserted into the placeholder $\textless\text{VAFeature}\textgreater$, while the {Text} section contains the tokenized textual input (Fig.~\ref{fig:promptset}).

Step 5: The fused audio-visual token $\widehat{VA}$ is inserted into the token sequence of task prompts and text tokens $\widehat{T}$, creating a structurally coherent and semantically aligned multimodal input. This sequence is then processed by the LLM for joint modeling.

Step 6: Minimize the discrepancy between predicted tokens $\widehat{t}_p$ and ground truth tokens $\widehat{t}_g$ using a similarity-based loss function (Eq. (\ref{eq:Lora})). LoRA is applied to fine-tune the LLM, improving its adaptability and discriminative ability in multimodal settings, thereby enhancing depression recognition performance.
\begin{equation}
\mathcal{L}_{LoRA}=\mathcal{L}_{ce}(\ \widehat{t}_p, \ \widehat{t}_g),
\label{eq:Lora}
\end{equation}
where $\mathcal{L}_{ce}(.,.)$ denotes the cross-entropy loss.

\section{Experimental setup}
\label{Experimental set}
\subsection{Dataset}
We evaluate our method on five publicly available multimodal depression datasets: CMDC \cite{CMDC}, AVEC2014 \cite{AVEC2014}, DAIC-WOZ \cite{DAIC}, DVlog \cite{DVlog}, and EATD \cite{EATD}, which differ in data sources and modality composition (see Table~\ref{table:contrast}).

\textbf{CMDC.} A Chinese multimodal dataset with 78 participants, including 19 depressed and 59 non-depressed samples, collected through semi-structured clinical interviews. It contains audio, text, and partial video features, with only 45 samples having all three modalities.

\textbf{AVEC2014.} A German dataset with 150 Freeform interview samples. Audio is extracted using ffmpeg and transcribed with iFlytek ASR, while visual features are extracted using OpenFace \cite{openface2.0}.

\textbf{DAIC-WOZ.} Contains 193 semi-structured clinical interviews with audio, text, and facial video data, with OpenFace features pre-extracted.

\textbf{D-Vlog.} A dataset of 961 video blogs from 816 unique YouTube users, with 555 labeled depressive and 406 non-depressive. Audio and visual features are extracted using the same pipeline as AVEC2014.

\textbf{EATD.} A dataset of 162 audio-text pairs from student volunteers, with 30 depressive and 132 non-depressive samples.

We unify the semantic space by translating German and Chinese transcripts to English using DeepL translator.

\subsection{Data Preprocessing and Augmentation}
\label{sec:dataprocess}
To improve model stability and generalization, we implement a preprocessing and augmentation pipeline addressing small sample sizes, modality heterogeneity, and class imbalance.
\begin{itemize}
    \item \textbf{Modality Parameter Unification}. To ensure consistency across datasets with different modality parameters, we apply a unified resampling strategy: all audio is downsampled to 16kHz mono, and video is reduced to 30fps. Additionally, visual frame sequences are temporally downsampled by a factor of 6 to preserve key dynamic features while reducing redundancy.
    \item \textbf{Long Sequence Samples Process}. We segment long sequences from the D-Vlog and DAIC-WOZ datasets using a 3-minute window to maximally comply with the input length limitations of Mistral-7B LLM \cite{jiang2023mistral7b}. Corresponding audio and video are trimmed synchronously to maintain alignment. Each segment is treated as an independent training sample, effectively expanding the dataset.
    \item \textbf{QA-Based Sample Augmentation}. To address small sample in short-sequence datasets like CMDC and EATD, we propose a QA-based augmentation strategy. Each QA (question–answer) pair is treated as an independent training instance, enriching semantic diversity and expanding the sample size. For example, each EATD sample with 3 QA pairs generates 3 instances, while CMDC samples with 12 QA pairs yield 12 instances. To balance class distribution, we introduce a QA recombination mechanism. It combines different QA pairs from depressive samples to form new QA pairs, which are treated as independent instances, thereby increasing the number of depressive samples and achieving class balance. This recombination mechanism is applied only to the official training split of CMDC and EATD, ensuring that the test split remains unchanged.
\end{itemize}

\subsection{Implementation Details}
All experiments were conducted on NVIDIA 4×A800 GPUs. We adopt Mistral-7B \cite{jiang2023mistral7b} as the backbone LLM due to its superior performance in our preliminary experiments, outperforming LLaMA 2-7B in training speed and validation accuracy. The model is trained on all cross-scenario multimodal depression datasets. During both training and inference, we use fixed LLM decoding configurations (top-p = 0.9, temperature = 1.0).

\subsubsection{Training Setup}
For depression classification, the model is trained for 400,000 steps on the merged dataset, using a global batch size of 8 and a maximum learning rate of 1e-5. LoRA hyperparameters are set to rank = 64, alpha = 16, and dropout = 0.05. For pretraining the audio, video encoders, and Video-Audio Fusion Module (VAFM), we use the Adam optimizer with a learning rate of 1e-4 for 100 epochs and a batch size of 4.

\subsubsection{Evaluation Protocol}
We use the official train-test splits provided by each dataset. Performance is evaluated by comparing the model's generated responses against ground-truth labels, calculating accuracy, precision, recall, and F1 score. If the generated response contains conflicting cues or lacks a clear affirmative statement, it is classified as "Error."

For evaluation, samples from the same participant are treated as independent instances, with a voting mechanism applied to the predictions. If more than half of the associated samples are classified as depressive, the participant is labeled as depressive; otherwise, non-depressive.

\section{Results and analysis}
\label{sec:results_analysis}
\subsection{Comparison of Varying Modality Inputs across Datasets}
\label{results:generation}
To evaluate the generalization ability of SCD-MLLM across different scenarios and modality combinations, we conducted experiments on five multimodal depression datasets: CMDC, AVEC2014, DAIC, DVlog, and EATD. The model was trained on all datasets together, and evaluated on each dataset using different modality combinations, including texts-only (T), texts with audios (T+A), texts with videos (T+V), and all modalities (T+A+V). As shown in Table~\ref{tab:multimodal}, SCD-MLLM achieves strong and consistent performance across datasets, highlighting its ability to generalize effectively across different scenarios and modality configurations.

Across datasets, multimodal fusion generally outperforms unimodal inputs, highlighting the complementary nature of textual, acoustic, and visual cues. Specifically, a clear gain from audio–text fusion is observed on \textbf{EATD}, where adding audio (T+A) lifts the F1-score from 0.6430 to 0.7492. A similar pattern appears on \textbf{AVEC2014} and \textbf{DVlog}, where the full T+A+V setting yields the best results among all modality combinations. In particular, on AVEC2014, T+A+V (F1 = 0.6940) outperforms text-only T (F1 = 0.4746) by about 22\% in F1 score, indicating that audio and video provide strongly complementary cues that substantially reinforce the textual signal.

Surprisingly, on \textbf{CMDC}, the model achieves perfect scores (accuracy and F1 = 1.0) across all modality combinations, indicating that textual responses alone provide sufficient depressive cues, and adding audio or visual features does not introduce conflicting information.

For \textbf{DAIC}, the model performs best with T+A (F1 = 0.6974), and adding video does not improve performance. This is likely because many interview questions, such as "Where is your hometown?", are unrelated to depression, causing visual features to introduce noise and not contribute meaningfully to depression recognition, as reflected in the T+V results.

Overall, SCD-MLLM excels in multimodal depression recognition by effectively leveraging cross-modal complementarity, particularly in scenarios with highly informative cues. The model demonstrates its capability to generalize across diverse datasets and modality combinations, making it adaptable for real-world, cross-scenario applications.

\begin{table}[!t]
\centering
\caption{Results of multimodal depression recognition across multiple datasets and modalities. This table presents the performance of SCD-MLLM on five depression datasets under various modality combinations. The results confirm the effectiveness of SCD-MLLM in generalization across heterogeneous datasets and in varying modality configurations.}
\footnotesize
\renewcommand{\arraystretch}{1.2}
\setlength{\tabcolsep}{2.1mm}
\begin{tabular}{llcccc}
\toprule
\textbf{Datasets} & \textbf{Modal} & \textbf{Accuracy} & \textbf{Precision} & \textbf{Recall} & \textbf{F1-score} \\
\midrule
\multirow{4}{*}{\textbf{CMDC}} 
    & T        & 1.0000     & 1.0000     & 1.0000     & 1.0000     \\
    & T+A      & 1.0000     & 1.0000     & 1.0000     & 1.0000     \\
    & T+V      & 1.0000    & 1.0000     & 1.0000     & 1.0000     \\
    & T+A+V    & \textbf{1.0000}     & \textbf{1.0000}     & \textbf{1.0000}     & \textbf{1.0000}     \\
\midrule
\multirow{2}{*}{\textbf{EATD}} 
    & T        & \underline{0.8354}  & \underline{0.6493}  & \underline{0.6377}  & \underline{0.6430} \\
    & T+A      & \textbf{0.8987}  & \textbf{0.8155}  & \textbf{0.7126}  & \textbf{0.7492} \\
\midrule
\multirow{4}{*}{\textbf{AVEC2014}} 
    & T        & 0.5306  & 0.5403  & 0.5242  & 0.4746 \\
    & T+A      & \underline{0.6939}  & \textbf{0.7235}  & \underline{0.6900}  & \underline{0.6806} \\
    & T+V      & 0.6000  & 0.6694  & 0.6000  & 0.5544 \\
    & T+A+V    & \textbf{0.7000}  & \underline{0.7170}  & \textbf{0.7000}  & \textbf{0.6940} \\
\midrule
\multirow{4}{*}{\textbf{DAIC}} 
    & T        & 0.7174  & 0.6355  & \textbf{0.6750}  & \underline{0.6430} \\
    & T+A      & \textbf{0.8261}  & 0.7583  & \underline{0.6722}  & \textbf{0.6974} \\
    & T+V      & 0.7955  & \underline{0.7358}  & 0.5853  & 0.5938 \\
    & T+A+V    & \underline{0.8182}  & \textbf{0.9048}  & 0.6000  & 0.6140 \\
\midrule
\multirow{4}{*}{\textbf{DVlog}} 
    & T        & 0.8895  & 0.8895  & 0.8930  & 0.8892 \\
    & T+A      & 0.9368  & 0.9364  & 0.9403  & 0.9367 \\
    & T+V      & \underline{0.9684}  & \underline{0.9681}  & \underline{0.9681}  & \underline{0.9681} \\
    & T+A+V    & \textbf{0.9737}  & \textbf{0.9730}  & \textbf{0.9739}  & \textbf{0.9735} \\
\bottomrule
\end{tabular}
\begin{flushleft}
{\footnotesize Note: T = Text, A = Audio, V = Video. Bold value = best, Underlined value= second-best.}
\end{flushleft}
\label{tab:multimodal}
\end{table}

\subsection{Comparison Under Missing Modalities}
\label{results:stability}
\begin{table}[!t]
\centering
\caption{Results of comparing the unified SCD-MLLM model under missing-modality inputs across three depression scenarios with dataset-specific models for each modality–dataset combination. All results are reported in terms of F1-score; overall, the general model (SCD-MLLM-G) matches or surpasses the specialized models (SCD-MLLM-S) under partial-modality settings, indicating stable performance with missing modalities and clear benefits from cross-dataset training.}
\footnotesize
\renewcommand{\arraystretch}{1.25}
\setlength{\tabcolsep}{2.6mm}
\begin{tabular}{c|c|c c c}
\hline
\multirow{2}{*}{\textbf{Modal}} & \multirow{2}{*}{\textbf{Model}} & \multicolumn{3}{c}{\textbf{F1-Score}} \\
\cline{3-5}
 & & \textbf{CMDC} & \textbf{DVLog} & \textbf{EATD} \\
\hline
\multirow{2}{*}{T} & SCD-MLLM-S & 0.9345 & 0.8891 & 0.5715 \\
  & SCD-MLLM-G (Ours) & \textbf{1} & \textbf{0.9050} & \textbf{0.6081} \\
\hline
\multirow{2}{*}{T+A} & SCD-MLLM-S & 0.9345 & 0.9157 & 0.6491 \\
     & SCD-MLLM-G (Ours) & 0.9345 & 0.9157 & \textbf{0.7492} \\
\hline
\multirow{2}{*}{T+V} & SCD-MLLM-S & 1 & 0.9017 & - \\
     & SCD-MLLM-G (Ours) & 1 & \textbf{0.9156} & - \\
\hline
\multirow{2}{*}{T+A+V} & SCD-MLLM-S & 1 & \textbf{0.9841} & - \\
      & SCD-MLLM-G (Ours) & 1 & 0.9735 & - \\
\hline
\end{tabular}
\label{tab:missing_modality}

\begin{flushleft}
{\footnotesize Note: T = Text, A = Audio, V = Video. Bold value is the best in respective modal configurations.}
\end{flushleft}
\end{table}

To examine the stability of our unified model when modalities are missing, we compare two variants: SCD-MLLM-G, a general model trained once on all datasets with T+A+V inputs, as in Sec.~\ref{results:generation}, and tested with the corresponding available modalities (T, T+A, T+V, T+A+V); and SCD-MLLM-S, a specialized model trained and evaluated on each dataset separately using the same modality configuration. Table~\ref{tab:missing_modality} reports F1-scores on three representative scenarios: CMDC (interview-based), DVlog (self-narration), and EATD (online questionnaires). Overall, SCD-MLLM-G matches or surpasses SCD-MLLM-S for all partial-modality settings, This indicates that a single cross-dataset T+A+V model can be used directly under missing-modality inputs with often outperforming dataset-specific models.

For all partial-modality settings (T, T+A, T+V), the unified model SCD-MLLM-G matches or surpasses the dataset-specific SCD-MLLM-S on CMDC, DVlog, and EATD, such as improving F1 from 0.9345 to 1.0000 on CMDC (T) and from 0.6491 to 0.7492 on EATD (T+A). This shows that training once on all datasets with full T+A+V inputs yields a generic model that remains reliable even when audio or video is missing at inference across all scenarios.

The only exception is the full T+A+V setting on DVlog, where SCD-MLLM-G (F1 = 0.9735) is slightly below the specialized model (F1 = 0.9841). This pattern highlights the strong capacity of the SCD-MLLM framework to exploit semantically rich self-narration data. Although the general model is slightly lower than the dataset-specific one, it retains competitive peak performance while offering broader applicability under missing-modality conditions.

In summary, these results show that SCD-MLLM, when trained once with full multimodal inputs across datasets, produces a robust generic model that tolerates missing modalities at inference, often outperforms modality- and dataset-specific baselines, and thus holds strong potential as a unified depression recognizer in realistic, modality-incomplete scenarios.

\subsection{Comparison with State-of-the-Art}
\label{results:sota}
\begin{table*}[!t]
\centering
\footnotesize
\caption{Results of performance comparison with state-of-the-art methods on datasets across three depression recognition scenarios: (a) Interview (CMDC), (b) Questionnaire (EATD), and (c) Self-narration (DVlog). All results are reported in terms of F1-score. Across all datasets of different source, SCD-MLLM achieves the best or competitive results, validating the effectiveness of our unified framework in adapting to diverse depression recognition scenarios and underscoring its potential for real-world deployment.}
\renewcommand{\arraystretch}{1.2}

\begin{minipage}{0.3\textwidth}
\centering
\setlength{\tabcolsep}{1.5mm}
\textbf{(a) CMDC}\\[4pt]
\begin{tabular}{l|c|c}
\hline
\textbf{Method} & \textbf{Modal} & \textbf{F1} \\
\hline
IISFD (2024) \cite{li2024enhancing}                & A+V     & 0.7800 \\
IIFDD (2024) \cite{IIFDD}                 & T+A     & 0.9500 \\
MDDMamba (2024) \cite{liu2024mddmamba}             & T+A     & 0.9700 \\
BMBRT (2024) \cite{jia2024bidirectional}           & T+A     & \underline{0.9800} \\
MLlm-DR (2025) \cite{zhang2025mllmdrexplainabledepressionrecognition} & T+A+V & \textbf{1.0000} \\
\cline{1-3}
SCD-MLLM (Ours)       & T+A+V   & \textbf{1.0000} \\
\hline
\end{tabular}
\end{minipage}
\hfill
\begin{minipage}{0.3\textwidth}
\centering
\setlength{\tabcolsep}{1.5mm}
\textbf{(b) DVlog}\\[4pt]
\begin{tabular}{l|c|c}
\hline
\textbf{Method} & \textbf{Modal} & \textbf{F1} \\
\hline
FairReFuse (2024) \cite{cheong2024fairrefuse}          & A       & 0.7000 \\
STAT (2024) \cite{tao2024depmstat}                     & A+V     & 0.7351 \\
STST (2024) \cite{tao2024depressive}                   & A+V     & 0.7500 \\
EMO-Mamba (2024) \cite{xing2024emo}                    & A+V     & \underline{0.7566} \\
DepMamba (2025) \cite{ye2025depmamba}                        & A+V     & 0.7512 \\
\cline{1-3}
SCD-MLLM (Ours)      & T+A+V   & \textbf{0.9735} \\
\hline
\end{tabular}
\end{minipage}
\hfill
\begin{minipage}{0.35\textwidth}
\centering
\setlength{\tabcolsep}{1.5mm}
\textbf{(c) EATD}\\[4pt]
\begin{tabular}{l|c|c}
\hline
\textbf{Method} & \textbf{Modal} & \textbf{F1} \\
\hline
GRU-BiLSTM(2022) \cite{EATD} & T+A     & 0.7100 \\
ROBERTa-BiLSTM(2022) \cite{zhang2022hybrid} & T     & 0.6900 \\
TAMFN (2023) \cite{zhou2022tamfn}              & T+A   & \textbf{0.7500} \\
IIFDD (2024) \cite{IIFDD}              & T+A   & 0.5000 \\
APAM (2025) \cite{goncc2025affect}              & T     & 0.7000 \\
\cline{1-3}
SCD-MLLM (Ours)     & T+A   & \underline{0.7492} \\
\hline
\end{tabular}
\end{minipage}

\begin{flushleft}
{\footnotesize Note: T = Text, A = Audio, V = Video. Bold value = best, Underlined value = second-best.}
\end{flushleft}
\label{tab:sota_all}
\end{table*}

To further assess the effectiveness of SCD-MLLM, we compare it with state-of-the-art depression recognition methods on three representative datasets that cover distinct use scenarios: CMDC (interview-based), DVlog (self-narration), and EATD (online questionnaires). All methods are evaluated using the F1-score, and the detailed results are reported in Table~\ref{tab:sota_all}. Overall, SCD-MLLM achieves the best or highly competitive performance on all three datasets, indicating strong robustness across tasks and data sources.

\textbf{CMDC}. On the interview-based CMDC dataset (Table~\ref{tab:sota_all} (a)), SCD-MLLM attains a perfect F1-score of 1.0, matching the best result of MLlm-DR~\cite{zhang2025mllmdrexplainabledepressionrecognition}. Unlike MLlm-DR, which is specifically tailored to interview data, SCD-MLLM is a unified architecture that can be directly applied to other scenarios without modification. 

\textbf{DVlog}. For the self-narration DVlog dataset (Table~\ref{tab:sota_all} (b)), SCD-MLLM achieves an F1-score of 0.9735, substantially outperforming previous approaches such as EMO-Mamba (0.7566) and STST (0.7500), reflecting the advantage of our modality-aware fusion in rich, spontaneous speech settings. 

\textbf{EATD}. On the questionnaire-based EATD dataset (Table~\ref{tab:sota_all} (c)), SCD-MLLM obtains an F1-score of 0.7492, which is nearly identical to the best-performing TAMFN model (0.7500), even though TAMFN is a dataset-specific transformer designed for this benchmark.

In summary, these results demonstrate that SCD-MLLM not only matches specialized methods on their target datasets but also delivers substantial gains on more challenging self-narration data, all within a single unified framework. The ability to remain competitive across heterogeneous datasets and scenarios, without dataset-specific tailoring, highlights the effectiveness of our multimodal fusion strategy and the practical potential of SCD-MLLM for real-world depression recognition applications.

\subsection{Comparison with Advanced Multimodal LLMs} 
\label{results:llm}
\begin{table}[!t]
\centering
\footnotesize
\caption{Results of performance comparison with advanced commercial LLMs (Gemini 2.5 and GPT-4o) across different modality settings on five depression datasets. All results are reported in terms of F1-score. Compared with commercial LLMs, Our SCD-MLLM achieves more stable and consistent performance across different datasets and modality combinations, demonstrating its superiority as a domain-adaptive LLM for depression recognition.}
\renewcommand{\arraystretch}{1.2}
\setlength{\tabcolsep}{1.6mm}
\begin{tabular}{llccccc}
\toprule
\multirow{2}{*}{\textbf{Modal}} & \multirow{2}{*}{\textbf{LLMs}} & \multicolumn{5}{c}{\textbf{Datasets}} \\
\cmidrule(lr){3-7}
& & \textbf{CMDC} & \textbf{EATD} & \textbf{AVEC} & \textbf{DAIC} & \textbf{DVlog} \\
\midrule
\multirow{3}{*}{T} 
    & Gemini 2.5       & \textbf{1.0000} & \textbf{0.7201} & 0.2192 & \underline{0.6071} & \underline{0.9000} \\
    & GPT-4o           & \textbf{1.0000} & 0.5611             & \underline{0.2698} & 0.5165 & \textbf{0.9156} \\
    & SCD-MLLM    & \textbf{1.0000} & \underline{0.6430}              & \textbf{0.4746} & \textbf{0.643} & 0.8892 \\
\midrule
\multirow{3}{*}{T+A} 
    & Gemini 2.5       & \textbf{1.0000} & \underline{0.7138}             & \underline{0.4432} & \underline{0.6763} & 0.6051 \\
    & GPT-4o           & \textbf{1.0000} & 0.643              & 0.3718 & 0.6071 & \underline{0.9103} \\
    & SCD-MLLM    & \textbf{1.0000} & \textbf{0.7492}    & \textbf{0.6806} & \textbf{0.6974} & \textbf{0.9367} \\
\midrule
\multirow{3}{*}{T+V} 
    & Gemini 2.5       & \textbf{1.0000} & -                  & \underline{0.4121} & 0.5879 & 0.6078 \\
    & GPT-4o           & \textbf{1.0000} & -                  & 0.2698 & \textbf{0.646} & \underline{0.9149} \\
    & SCD-MLLM    & \textbf{1.0000} & -                  & \textbf{0.5544} & \underline{0.5938} & \textbf{0.9681} \\
\midrule
\multirow{3}{*}{T+A+V} 
    & Gemini 2.5       & 0.8831 & - & \underline{0.2756} & 0.6071 & 0.5994 \\
    & GPT-4o           & \underline{0.9345} & - & 0.2619 & 0.5368 & \underline{0.9367} \\
    & SCD-MLLM    & \textbf{1.0000} & - & \textbf{0.6566} & \textbf{0.614} & \textbf{0.9735} \\
\bottomrule
\end{tabular}
\begin{flushleft}
{\footnotesize Note: T = Text, A = Audio, V = Video. Bold value = best, Underlined value = second-best.}
\end{flushleft}
\label{tab:gpt_gemini}
\end{table}

Large language models (LLMs) such as Gemini 2.5 and GPT-4o have strong text understanding and emerging multimodal capabilities. To assess whether a domain-adaptive LLM brings additional benefits, we compare SCD-MLLM with these commercial models on five depression datasets and four modality settings (T, T+A, T+V, T+A+V). Since Gemini 2.5 and GPT-4o do not natively accept time-series features from OpenFace, we convert the facial-landmark frame sequences into short video clips at 30 frames per second and use them as visual inputs. The F1-scores are reported in Table~\ref{tab:gpt_gemini}.

Across most datasets and modality configurations, SCD-MLLM attains the best or second-best F1-score and shows fewer extreme failures than the commercial baselines. Under the full T+A+V setting, it clearly outperforms Gemini 2.5 and GPT-4o on AVEC2014 and DVlog (e.g., 0.6566 vs.\ 0.2756 on AVEC2014, and 0.9735 vs.\ 0.9367 on DVlog), indicating stronger adaptation to complex multimodal signals. Even in the text-only setting on AVEC2014, SCD-MLLM improves F1 by more than 25\% over Gemini 2.5.

These comparisons show that a carefully designed, domain-adaptive multimodal LLM can match or exceed the performance of much larger general-purpose LLMs on depression recognition, while providing more stable behavior across datasets and modality settings.

\subsection{Ablations Studies}
\subsubsection{Comparison of Video-Audio Fusion Strategies}
\label{results:va_fusion}
\begin{table}[!t]
\centering
\caption{Results of comparison of V–A fusion methods across datasets. All models are evaluated under consistent T+A+V settings. Results demonstrate that Our SCD-MLLM outperforms or matches baselines, highlighting its effectiveness in capturing multimodal complementarity for generating semantically rich representations.}
\footnotesize
\renewcommand{\arraystretch}{1.2}
\setlength{\tabcolsep}{1.4mm}
\begin{tabular}{lccccc}
\toprule
\textbf{Datasets} &\textbf{Fusion Methods} & \textbf{Accuracy} & \textbf{Precision} & \textbf{Recall} & \textbf{F1} \\
\midrule
\multirow{3}{*}{\textbf{CMDC}} 
    & T-MLLM           & \underline{0.9444} & \underline{0.9615} & \underline{0.9167} & \underline{0.9345} \\
    & AffectGPT        & 0.8889 & 0.9286 & 0.8333 & 0.8615 \\
    & SCD-MLLM    & \textbf{1.0000} & \textbf{1.0000} & \textbf{1.0000} & \textbf{1.0000} \\
\midrule
\multirow{3}{*}{\textbf{AVEC2014}} 
    & T-MLLM           & 0.6000 & 0.6299 & 0.6000 & 0.5756 \\
    & AffectGPT        & \underline{0.6800} & \textbf{0.7339} & \underline{0.6800} & \underline{0.6604} \\
    & SCD-MLLM    & \textbf{0.7000} & \underline{0.7170} & \textbf{0.7000} & \textbf{0.6940} \\
\midrule
\multirow{3}{*}{\textbf{DAIC}} 
    & T-MLLM           & 0.7955 & 0.7046 & \textbf{0.6559} & \textbf{0.6719} \\
    & AffectGPT        & \underline{0.7955} & \underline{0.7103} & \underline{0.6206} & \underline{0.6384} \\
    & SCD-MLLM    & \textbf{0.8182} & \textbf{0.9048} & 0.6000 & 0.6140 \\
\midrule
\multirow{3}{*}{\textbf{DVlog}} 
    & T-MLLM           & 0.9316 & 0.9328 & 0.9365 & 0.9315 \\
    & AffectGPT        & \underline{0.9368} & \underline{0.9364} & \underline{0.9403} & \underline{0.9367} \\
    & SCD-MLLM    & \textbf{0.9737} & \textbf{0.9730} & \textbf{0.9739} & \textbf{0.9735} \\
\bottomrule
\end{tabular}
\begin{flushleft}
{\footnotesize Note: T = Text, A = Audio, V = Video. Bold value = best, Underlined value = second-best.}
\end{flushleft}
\label{tab:va_fusion}
\end{table}

To evaluate the effectiveness of the proposed MAFM, which employs a unified fusion module and a single shared linear projection to align audio and visual features in the LLM embedding space, we compare SCD-MLLM with two representative fusion baselines: (i) a traditional multimodal LLM pipeline (T-MLLM) \cite{cheng2024emotion} that uses separate linear projections for audio and video, and (ii) AffectGPT \cite{lian2025affectgpt}, which augments unimodal embeddings with an additional fused A+V representation. All models are evaluated under the same T+A+V setting on four datasets with paired audio–video data: CMDC, AVEC2014, DAIC, and DVlog. EATD is excluded due to the lack of video. As reported in Table~\ref{tab:va_fusion}, SCD-MLLM achieves the best or highly competitive results on most datasets, indicating that MAFM is an effective and parameter-efficient fusion strategy.

Across \textbf{CMDC}, \textbf{AVEC2014}, and \textbf{DVlog}, SCD-MLLM consistently achieves the best F1-scores among the three fusion strategies, highlighting the benefit of our modality-aware fusion with a shared projection. For example, on CMDC, SCD-MLLM improves F1 from 0.9345 (T-MLLM) to 1.0000; on AVEC2014, it raises F1 from 0.6604 (AffectGPT) to 0.6940; and on DVlog, it boosts F1 from 0.9367 (AffectGPT) to 0.9735. These consistent gains show that our fusion design more effectively exploits complementary audio and video cues than both separate linear mappings (T-MLLM) and the additional fused A+V branch in AffectGPT.

Specifically, on \textbf{DAIC}, SCD-MLLM reaches the highest accuracy (0.8182) but a slightly lower F1-score (0.6140), reflecting high precision but reduced recall. DAIC includes many non-depression-related interview segments and weak emotional alignment across modalities; in this case, stronger fusion may mix informative and noisy cues, whereas the more separated representations in T-MLLM sometimes better preserve distinct unimodal signals.

Overall, these comparisons show that SCD-MLLM with MAFM provides a strong and generally superior video–audio fusion mechanism. It delivers clear improvements on semantically coherent, low-resource datasets, remains competitive on more challenging benchmarks, and offers a unified, parameter-efficient design that naturally accommodates missing-modality cases.

\begin{figure}[!t]
\centering
\includegraphics[width=0.48\textwidth]{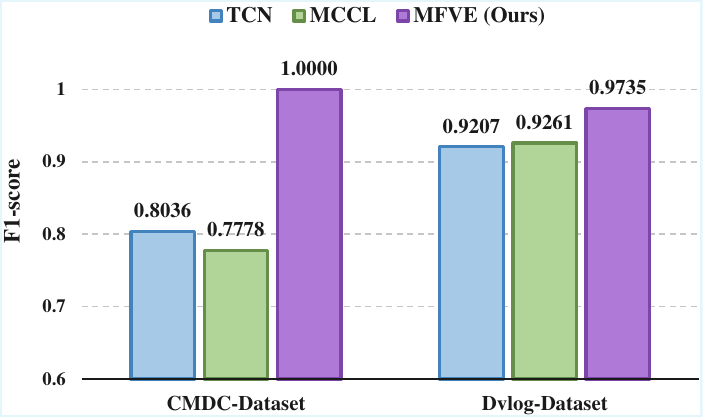}
\caption{Results of comparison of different video extraction methods. All comparisons are conducted under the T+A+V modality configuration, and F1-score is used as the performance metric. The results demonstrate the superiority of our method over existing fusion strategies, highlighting its more effective integration of heterogeneous visual cues.} 
\label{fig:VideoExtraction_Comparison}
\vspace{-0.2cm}
\end{figure}

\subsubsection{Comparison with Video Extraction Method}
\label{results:videoEncoder}
To make full use of visual cues extracted by OpenFace, SCD-MLLM employs the proposed Multi-Cue Fusion Video Encoder (MFVE), which fuses four heterogeneous facial signals (landmarks, gaze, head pose, and action units) into a unified video embedding aligned with the audio and text representations.

We evaluate MFVE on the CMDC and DVlog datasets by comparing it with two representative fusion strategies: (i) the Temporal Convolutional Network (TCN)–based fusion used in LMVD~\cite{LMVD}, and (ii) the Multi-Cue Contrastive Learning (MCCL) method~\cite{wang2025facial} that models inter-cue relations via attention. All methods are trained and evaluated under the same T+A+V setting, and performance is measured by F1-score (see Fig.~\ref{fig:VideoExtraction_Comparison}).

MFVE consistently outperforms the TCN and MCCL baselines on both datasets. On CMDC, for example, it improves F1 by 19.4\% and 22.2\% over TCN and MCCL, respectively, confirming the benefit of unified multi-cue encoding for depression-related visual signals.

We attribute these gains to MFVE’s ability to standardize heterogeneous cues and capture high-order interactions among them through cross- and self-attention, yielding more discriminative video-level embeddings. This stronger visual representation, in turn, enhances the overall multimodal understanding of SCD-MLLM for depression recognition.

\section{Conclusion}
\label{conclusion}
In this paper, we present SCD-MLLM, a novel multimodal large language model designed for depression recognition across diverse recognition scenarios, including clinical environments, online assessments, and self-reported emotional evaluations, and situations involving missing modalities. Our approach adopts Mistral 7B LLM as the language decoder, which is fine-tuned on prompt-based question-answering tasks specifically tailored for depression assessment. To facilitate the effective integration of heterogeneous inputs, we propose a modality-aligned tokenization pipeline that transforms diverse textual structures, variable-length audio signals, and semantically distinct visual streams into fixed-length token sequences within the embedding space of LLM. Additionally, we introduce a modality-aware adaptive fusion module to enhance the representational quality of audio and visual cues while adaptively managing missing modalities. Experiments results confirm the effectiveness of our proposed framework. SCD-MLLM surpasses strong baseline models and leading commercial LLMs under both complete and partial modality settings, demonstrating superior cross-domain generalization, the ability to harness multimodal emotional cues, and adaptability to practical deployment scenarios. 


\section{References Section}
%

\bibliographystyle{IEEEtran}
\bibliography{ref}


\newpage

 




\vfill

\end{document}